\documentclass[11pt]{article}

\usepackage[preprint]{acl}

\usepackage{times}
\usepackage[T1]{fontenc}
\usepackage[utf8]{inputenc}
\usepackage{microtype}
\IfFileExists{inconsolata.sty}{\usepackage{inconsolata}}{}
\usepackage{graphicx}

\usepackage{amsmath,amssymb,amsthm}
\usepackage{booktabs}
\usepackage{colortbl,tabularx}
\usepackage{adjustbox}
\usepackage{enumitem}
\usepackage{caption}
\usepackage{float}
\usepackage{stfloats}
\usepackage{xspace}
\usepackage{pifont}
\usepackage{tcolorbox}
\usepackage[capitalize,noabbrev]{cleveref}

\newcommand{\ttrsi}{TT-RSI\xspace}                 
\newcommand{\ttrsilong}{Test-Time Recursive Self-Improvement\xspace}
\newcommand{\icrsi}{In-Context RSI\xspace}         
\newcommand{\iwrsi}{In-Weight RSI\xspace}          

\definecolor{oursblue}{HTML}{679BC4}
\definecolor{oursgain}{HTML}{245581}
\newcommand{\ourmethodrow}{\rowcolor{oursblue!22}}

\newcommand{\scoregain}[1]{{\normalfont\textcolor{oursgain}{(#1\%)}}}

\newcommand{\A}{\mathcal{A}}
\newcommand{\E}{\mathbb{E}}
\newcommand{\KL}{\mathrm{KL}}
\newcommand{\sg}{\operatorname{sg}}
\newcommand{\TV}[1]{\lVert #1 \rVert_{\mathrm{TV}}}
\newcommand{\acc}{\operatorname{acc}}
\newcommand{\one}{\mathbf{1}}                       
\newcommand{\pinf}{\pi_{\infty}}
\newcommand{\dob}{\delta}                          

\theoremstyle{plain}
\newtheorem{theorem}{Theorem}
\newtheorem{proposition}{Proposition}
\newtheorem{lemma}{Lemma}
\newtheorem{corollary}{Corollary}
\theoremstyle{definition}
\newtheorem{definition}{Definition}

\newfloat{algobox}{tbp}{loa}
\floatname{algobox}{Algorithm}
\crefname{equation}{Eq.}{Eqs.}
\Crefname{equation}{Eq.}{Eqs.}
\crefname{algobox}{Algorithm}{Algorithms}
\Crefname{algobox}{Algorithm}{Algorithms}
\crefname{proposition}{Proposition}{Propositions}
\crefname{corollary}{Corollary}{Corollaries}
\crefname{definition}{Definition}{Definitions}

\newtcolorbox{promptbox}[1][]{colback=gray!6,colframe=gray!50,boxrule=0.4pt,
  arc=1pt,left=3pt,right=3pt,top=2pt,bottom=2pt,fontupper=\footnotesize\ttfamily,#1}

\newcommand{\overviewfigure}{%
\noindent\begin{minipage}{\textwidth}
\centering
\includegraphics[width=0.9\textwidth]{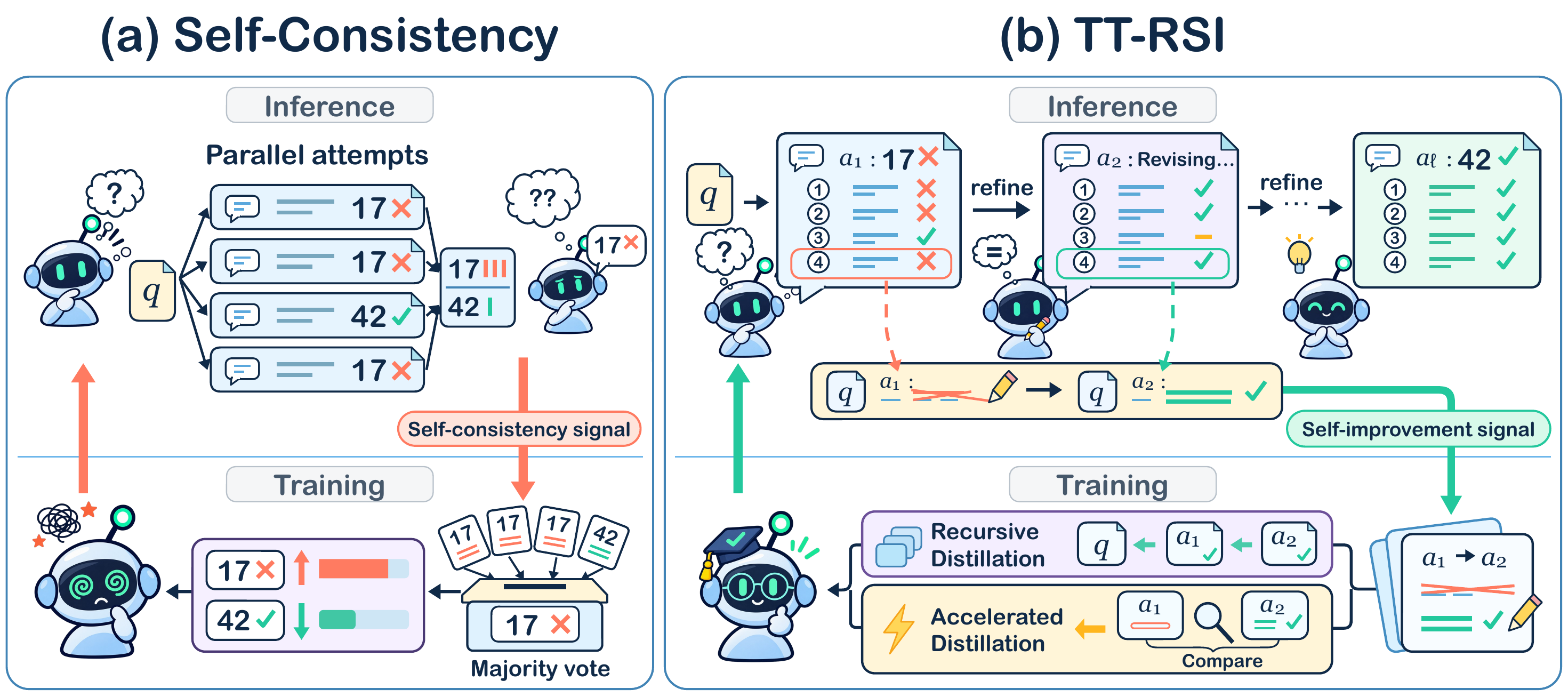}
\captionof{figure}{\textbf{Self-Consistency vs.\ \ttrsi{}.}
At test time, a model must judge its own answers on unlabeled questions, with no ground truth to tell which to trust.
(a)~Self-consistency-based methods resolve this by majority voting, taking the most frequent answer as the final output and as the training target, leaving prior reasoning unused at inference and reinforcing wrong answers during training.
(b)~\ttrsi reuses reasoning for efficient inference and trains on self-improvement to strengthen answering and refinement.}
\label{fig:overview}
\end{minipage}\par
}
\makeatletter
\newcommand{\maketitlewithteaser}{%
  \let\ttrsi@originalmaketitle\@maketitle
  \def\@maketitle{%
    \ttrsi@originalmaketitle
    \overviewfigure
    \vspace{\baselineskip}%
  }%
  \maketitle
}
\makeatother

\title{Beyond Self-Consistency: Recursive Self-Improvement at Test Time}
\hypersetup{pdftitle={Beyond Self-Consistency: Recursive Self-Improvement at Test Time},
  pdfauthor={Jianyu Wu, Yizhou Wang, Encheng Su, Jinouwen Zhang, Pengze Li, Lintao Wang, Chen Tang, Aoran Wang, Wanli Ouyang, Shixiang Tang}}

\author{%
  Jianyu Wu\textsuperscript{1,2}\quad
  Yizhou Wang\textsuperscript{2,3}\quad
  Encheng Su\textsuperscript{2,4}\quad
  Jinouwen Zhang\textsuperscript{2}\quad
  Pengze Li\textsuperscript{2,5} \\
  \bfseries Lintao Wang\textsuperscript{2,6}\quad
  Chen Tang\textsuperscript{2,3}\quad
  Aoran Wang\textsuperscript{2}\quad
  Wanli Ouyang\textsuperscript{3}\quad
  Shixiang Tang\textsuperscript{2,3,*} \\
  {\small\normalfont\textsuperscript{1}Shanghai Jiao Tong University\quad
   \textsuperscript{2}Shanghai Artificial Intelligence Laboratory} \\
  {\small\normalfont\textsuperscript{3}The Chinese University of Hong Kong\quad
   \textsuperscript{4}The University of Science and Technology of China} \\
  {\small\normalfont\textsuperscript{5}Fudan University\quad
   \textsuperscript{6}The University of Sydney\quad
   \textsuperscript{*}Corresponding author} \\
  {\small\texttt{wujianyu@sjtu.edu.cn}\quad
   \texttt{tangshixiang@pjlab.org.cn}}}

\begin{document}
\maketitlewithteaser

\begin{abstract}
Large language models (LLMs) that continue improving after deployment (e.g., recursive self-improvement) offer a promising path toward general intelligence.
Current test-time methods pursue this goal on unlabeled questions, primarily relying on self-consistency: they generate independent solutions and use majority voting to select an answer or construct training targets.
This leaves earlier reasoning unused at inference and can reinforce an incorrect majority during training.
Humans instead build on earlier attempts to preserve useful reasoning, correct mistakes and improve their problem-solving strategies.
Following this principle, we propose \ttrsilong (\ttrsi), a label-free framework for improving LLMs at test time through two mutually reinforcing components:
\textbf{(1)~\icrsi} repeatedly re-solves each question using the full previous solution as a reference, letting the model decide whether and how to use it so that inference compute builds on earlier reasoning.
\textbf{(2)~\iwrsi} distills successive refinements into the model's weights, enabling progress beyond an initially wrong majority and theoretically bringing the first attempt toward the model's infinite-step refinement limit.
The updated model answers better and refines more effectively, generating stronger references for subsequent training.
Extensive experiments show that \ttrsi improves average accuracy by \textbf{+13.89\%} over the base model and outperforms the strongest self-consistency-based method by \textbf{+4.79\%} under the same test-time scaling budget.
\end{abstract}

\section{Introduction}
\label{sec:intro}

Enabling AI systems to improve their own capabilities (e.g., recursive self-improvement \citep{novikov2025alphaevolve,zhang2026dgm} and continual learning \citep{ye2026oel}) is central to progress toward artificial general intelligence \citep{schmidhuber2007godel,silver2025experience}.
For LLMs, test time is a natural setting for such improvement:
after a model is deployed, the goal is to answer new, unlabeled
questions more accurately using only self-generated signals.
Two families of methods pursue this goal: \emph{test-time scaling} spends more inference compute on each question \citep{snell2025scaling,guo2025deepseekr1}, while \emph{test-time training} updates the model's weights on unlabeled test questions \citep{sun2020ttt,zuo2025ttrl}.
Ideally, the two form a loop: better answers train a better model, which in turn produces better answers, allowing improvements to accumulate over successive rounds without external supervision.

However, current test-time methods rely primarily on \emph{Self-Consistency} (SC) for both inference and training: they sample independent solutions and use majority voting either to select the final answer \citep{wang2023selfconsistency,brown2024monkeys} or to construct training targets \citep{zuo2025ttrl,wang2026ttpo,gkountouras2026canon}.
This leads to two linked limitations:
\textbf{(1) Inefficient inference.} Each attempt starts from scratch and leaves earlier reasoning unused, so additional compute yields more independent samples rather than better reasoning.
\textbf{(2) Unreliable supervision.} Majority voting rewards whichever answer the model already produces most often, so when that majority is wrong, training reinforces the error.
Together, these limitations break the loop above, so test-time improvement stalls rather than accumulates.

In contrast, humans tackle difficult problems by building on earlier attempts, retaining useful reasoning, revising mistakes, and gradually improving their problem-solving strategies \citep{polya1945solveit}.
Following this principle, we propose \ttrsilong (\ttrsi), a label-free framework that improves LLMs at test time through both inference and training (\cref{fig:overview}).
Both rely on an \emph{improvement step}, which re-solves a question with the previous solution as context. Specifically,
at inference, additional compute builds on earlier reasoning instead of repeating it.
During training, the model learns from its own refinements rather than majority votes.
Since a refinement can fix an earlier attempt even when most attempts agree on a wrong answer, this supervision can move the model beyond its initial majority,
yielding higher accuracy than majority-based training while
preserving the gains from further test-time scaling.

\begin{figure*}[t]
\centering
\includegraphics[width=\textwidth]{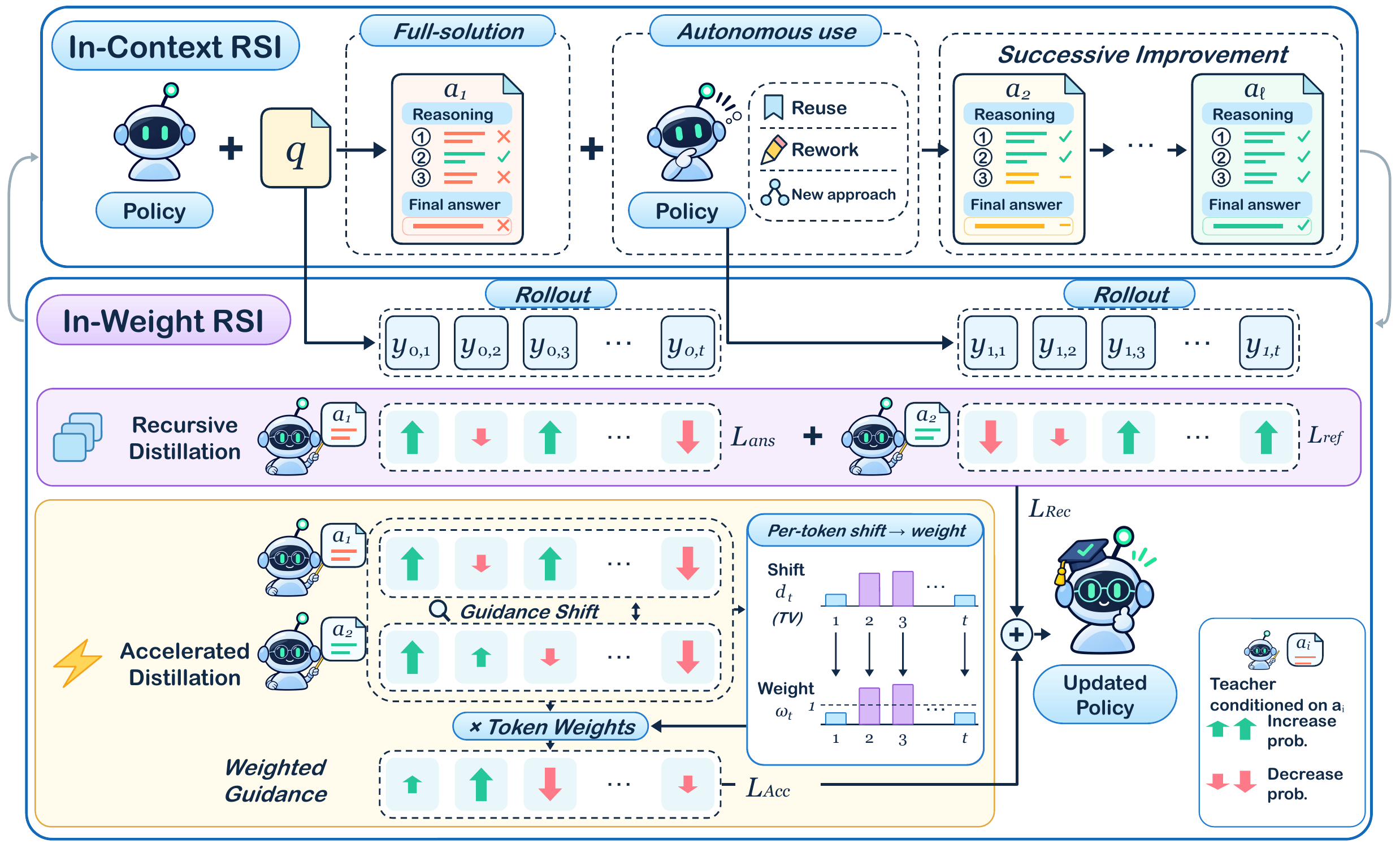}
\caption{\textbf{Overview of TT-RSI.}
\textbf{In-Context RSI} generates $a_1$ from $q$, then repeatedly re-solves $q$ given the full previous solution.
The model autonomously chooses what to reuse, rework, or solve anew, producing $a_2,\ldots,a_\ell$.
\textbf{In-Weight RSI} uses these solutions as teacher references: \textbf{Recursive Distillation} combines $a_1$- and $a_2$-conditioned guidance for answering ($L_{\mathrm{ans}}$) and refinement ($L_{\mathrm{ref}}$) as $L_{\mathrm{Rec}}$.
\textbf{Accelerated Distillation} compares the $a_1$- and $a_2$-conditioned guidance on $y_0$ and weights only $a_2$ guidance ($L_{\mathrm{Acc}}$), with larger shifts receiving larger weights.
The updated policy starts the next refinement chain.}
\label{fig:method}
\end{figure*}

\ttrsi realizes the improvement step through two complementary components.
At inference, we introduce \textbf{\icrsi}, a training-free test-time scaling strategy that turns additional compute into successive refinements.
Given a question, the LLM first generates an initial solution, then repeatedly re-solves the question with its latest solution as context, and returns the final answer of the last attempt.
However, simply conditioning on previous solutions can restrict exploration and cause repetition \citep{parseq2026gap,seqexplore2026}.
We therefore present the full solution neutrally, preserving its intermediate reasoning while letting the LLM decide whether and how to use it.
With the previous solution in view, the LLM need not re-derive every step: it can keep verified steps, focus on questionable ones, and avoid repeating earlier dead ends.
As a result, \icrsi surpasses self-consistency while reducing generated tokens by a relative \textbf{68.6\%} (\cref{fig:tts}).
These successive solutions also serve as training signals.

During training, we introduce \textbf{\iwrsi}, a label-free test-time training method that \emph{distills} the gains of \icrsi into the model's weights.
\iwrsi uses the first attempt and its immediate refinement, a pair of adjacent solutions from an \icrsi chain generated by the current policy, as references, strengthening both answering and refinement while transferring later improvements into the first attempt. Concretely, it learns from these references through on-policy self-distillation \citep[OPSD;][]{zhao2026opsd}, which turns them into token-level guidance on the model's own rollouts.
Because each refinement re-examines the full previous derivation, it can correct intermediate steps even when the final answer remains wrong.
Token-level guidance captures such partial corrections, which majority voting over final answers cannot.
In this way, \icrsi and \iwrsi reinforce each other: the refinements of \icrsi train a better model through \iwrsi, and the updated model refines more effectively under \icrsi, generating stronger references for later training.
This realizes the ideal loop between test-time scaling and training described above, so improvements can be accumulated without external supervision.

We validate \ttrsi with Qwen3-1.7B, 4B and 8B \citep{yang2025qwen3} on five competition mathematics benchmarks \citep{balunovic2025matharena,dekoninck2026matharena} against five label-free baselines.
\iwrsi and the full \ttrsi improve average accuracy over the base model by \textbf{+5.79\%} and \textbf{+13.89\%}, respectively.
Under the same test-time scaling budget, \ttrsi exceeds the strongest self-consistency-based method, TTPO + SC \citep{wang2026ttpo}, by \textbf{+4.79\%} on average.
Our contributions are as follows:
\begin{itemize}[leftmargin=*,itemsep=1pt,topsep=2pt]
\item \textbf{A recursive self-improvement framework.} We propose \ttrsi, which goes beyond self-consistency by using self-improvement for both test-time scaling and label-free test-time training.
\item \textbf{Two mutually reinforcing components.} We design \icrsi to let the model build on its full previous solution, and \iwrsi to distill successive refinements into answering and refinement, so that each strengthens the other.
\item \textbf{Theory and experiments.} We prove that distilling from two successive solutions strengthens refinement and moves the first attempt toward the infinite-step limit, and demonstrate accuracy and token-efficiency gains empirically.
\end{itemize}

\section{Preliminaries}
\label{sec:prelim}

\paragraph{Notation.}
Let $q$ be a question and $\pi_\theta$ the model.
We sample a visible solution $a\sim\pi_\theta(\cdot\mid q)$, consisting of a derivation and a final answer.

\paragraph{Label-free test-time training.}
Given the unlabeled questions of a test set, a test-time training method updates $\pi_\theta$ on those questions using only signals that the model generates itself \citep{sun2020ttt,zuo2025ttrl}.

\paragraph{On-policy self-distillation.}
OPSD \citep{zhao2026opsd} uses a frozen initial-model teacher $\pi_{\theta_0}$ with privileged information to guide $\pi_\theta$ on its own rollout.
For student context $x$ and teacher reference $y^\star$, the loss is
\begin{equation}
L_{\mathrm{OPSD}}=D\big(\sg[\pi_{\theta_0}(\cdot\mid x,y^\star)]\,\big\|\,\pi_\theta(\cdot\mid x)\big),
\label{eq:opsd}
\end{equation}
where $\sg$ denotes stop-gradient.
The teacher's next-token predictions provide guidance along a student rollout $y\sim\pi_\theta(\cdot\mid x)$, and $D$ compares them with the student's predictions at the same prefixes.

\section{Test-Time Recursive Self-Improvement}
\label{sec:method}

\ttrsi applies the improvement step (\cref{sec:operator}) in two components: \icrsi (\cref{sec:icrsi}) repeats it to refine answers at inference, while \iwrsi (\cref{sec:iwrsi}) turns the resulting refinements into supervision for answering and refinement (\cref{fig:method}).

\subsection{The improvement step}
\label{sec:operator}

At the core of \ttrsi is an \emph{improvement step}, which lets the model improve on its own attempts without labels by turning a completed attempt into context for further reasoning.
A previous solution provides a starting point from which the model can reuse useful parts of an argument, reconsider intermediate steps, and develop a better solution.
Starting from $a_1\sim\pi_\theta(\cdot\mid q)$, repeated application gives $a_{k+1}\sim\pi_\theta(\cdot\mid q,a_k)$.
Only the latest solution $a_k$ is carried forward, so the sequence forms a Markov chain without an expanding history.
Because later solutions tend to improve on earlier ones (\cref{fig:tts}), they can also supervise earlier attempts during training, distilling later improvements into the model (\cref{sec:iwrsi}).
Further details of the refinement setup are provided in \cref{app:prompts}.

\subsection{\icrsi}
\label{sec:icrsi}

Even an unsuccessful attempt often contains correct intermediate steps, yet independent sampling discards them and starts each attempt from scratch.
To put them to use, we propose \icrsi, a training-free test-time scaling strategy that applies the improvement step repeatedly to the same question, so that each attempt builds on the previous one.
Effective refinement needs information from an earlier attempt and the freedom to reconsider it.
Yet conditioning too strongly on an earlier solution can suppress exploration and lead to repeated solutions \citep{parseq2026gap,seqexplore2026}.
Directly applying correction-oriented feedback can also induce unnecessary changes to a correct answer \citep{huang2024cannot,kamoi2024correct}.

To address this tension, \icrsi combines full-solution references with autonomous reference use.
Given $q$, the policy first generates an initial solution $a_1\sim\pi_\theta(\cdot\mid q)$.
Each subsequent attempt then receives a \emph{full-solution reference}: the complete previous solution $a_k$, including both its step-by-step derivation and its final answer, rather than the final answer alone.
Attempt $a_{k+1}$ can thus check the individual steps of $a_k$, keep those that hold, and locate where the derivation went wrong, so even a solution with an incorrect final answer contributes reusable steps.
For \emph{autonomous reference use}, we present $a_k$ neutrally as ``a previous attempt'', without indicating whether it is correct or requesting corrections, and ask the policy to ``solve the problem independently'' (\cref{app:prompts}).
The policy itself decides how to use each part of $a_k$: reuse it, rework it, or take a new approach.
Repeating this step yields a chain $a_1\to a_2\to\cdots\to a_\ell$ of one initial solution and $\ell{-}1$ refinements, and we return the final answer of $a_\ell$.

To characterize the balance between correcting and preserving answers, we use a two-state model \citep{yang2025selfcorrectiontheory,liu2025dser,he2025instability}.
Each refinement step corrects a wrong answer with probability $f$, the \emph{fix rate}, and breaks a correct one with probability $b$, the \emph{break rate}.

\begin{proposition}
\label{prop:dyn}
For constant fix/break rates with $f{+}b\in(0,1]$, \icrsi improves its first-attempt accuracy $\acc_1$ exactly when $f\,(1-\acc_1)>b\,\acc_1$.
Its limiting accuracy, the \emph{balance point}, is $f/(f{+}b)$.
\end{proposition}

A refinement design should therefore raise the fix rate relative to the break rate.
We compare these rates for \icrsi and its design variants in \cref{sec:exp-ablation}.

\subsection{\iwrsi}
\label{sec:iwrsi}

The improvements of \icrsi last only for the current question, and majority-vote targets can reinforce a wrong answer.
We therefore propose \iwrsi, a label-free test-time training method that distills successive refinements into the model's weights.
Taking the first two solutions $a_1,a_2$ of an \icrsi chain generated by $\pi_\theta$ as references, \iwrsi trains answering and refinement through Recursive Distillation and couples them through Accelerated Distillation, optimizing both objectives jointly.
Both use OPSD (\cref{eq:opsd}), with these references as privileged information.

\paragraph{Recursive Distillation.}
A key design question is which solution along the refinement chain should supervise which attempt.
Ideally, training would transfer the result of infinitely many refinements directly into the first attempt, absorbing the full benefit of the chain.
However, long chains are costly to generate, and their late solutions can differ substantially from the student's early attempts, making the teacher's guidance on the student's own rollouts less reliable. This raises a question: is it possible to supervise only local relationships along the chain and thereby avoid these issues?

Fortunately, the chain involves only two behaviors: answering the question and refining the previous attempt.
Every later step reuses the refinement operation, so training these two behaviors covers the recurring roles along the chain.
For each behavior, we sample a student rollout: $y_0\sim\pi_\theta(\cdot\mid q)$ for answering and $y_1\sim\pi_\theta(\cdot\mid q,a_1)$ for refinement.
The teacher receives $a_1$ for answering and $a_2$ in place of $a_1$ for refinement:
\begin{equation}
\begin{aligned}
L_{\mathrm{ans}}&=D\big(\sg[\pi_{\theta_0}(\cdot\mid q,a_1)]\,\|\,\pi_\theta(\cdot\mid q)\big),\\
L_{\mathrm{ref}}&=D\big(\sg[\pi_{\theta_0}(\cdot\mid q,a_2)]\,\|\,\pi_\theta(\cdot\mid q,a_1)\big).
\end{aligned}
\label{eq:adjacent}
\end{equation}
The losses are evaluated on $y_0$ and $y_1$, respectively.
We call $L_{\mathrm{ans}}$ \emph{answer distillation}: its teacher has already seen $a_1$, so it transfers one improvement step into the first attempt.
We call $L_{\mathrm{ref}}$ \emph{refinement distillation}: its teacher has seen the refined solution $a_2$, so it trains the improvement step to incorporate a further refinement.
We combine them as
\begin{equation}
L_{\mathrm{Rec}}=L_{\mathrm{ans}}+L_{\mathrm{ref}}.
\label{eq:recursive}
\end{equation}
Although $L_{\mathrm{Rec}}$ needs only two reference solutions per question, it can recover the effect of much longer chains over repeated updates.
After each update, both behaviors improve, so the references $a_1,a_2$ generated by the updated policy correspond to later attempts of the original chain.
Distilling them again moves the first attempt and each refinement step further along that chain.
Each update thus needs only one-step-ahead targets, while the gains compound across updates (\cref{thm:main}).

\paragraph{Accelerated Distillation.}
$L_{\mathrm{ans}}$ and $L_{\mathrm{ref}}$ remain loosely coupled: the first attempt does not receive direct supervision from the later reference $a_2$.
Accelerated Distillation connects them by supervising the first-attempt rollout $y_0$ with the $a_2$-conditioned teacher of refinement distillation:
\begin{equation}
L_{\mathrm{Acc}}=D_\omega\big(\sg[\pi_{\theta_0}(\cdot\mid q,a_2)]\,\|\,\pi_\theta(\cdot\mid q)\big).
\label{eq:accelerated}
\end{equation}
Because $L_{\mathrm{ans}}$ already supplies $a_1$-conditioned guidance, we emphasize what changes when the reference is refined to $a_2$.
We compare the teacher's next-token distributions under $a_1$ and $a_2$ at the same prefix $y_{0,<t}$, and define the \emph{guidance shift} $d_t$ from $a_1$ to $a_2$ as the total-variation distance between them.
It determines the \emph{token weight}
\begin{equation}
\omega_t=1-\alpha+\frac{2\alpha\,d_t}{d_t+\tau},
\label{eq:omega}
\end{equation}
where $0<\alpha<1$ controls the weighting strength and $\tau>0$ sets the change scale.
$D_\omega$ multiplies each token's distillation loss on $y_0$ by $\omega_t$.
Larger shifts strengthen the $a_2$-conditioned supervision at positions affected by refinement.

In total, \iwrsi minimizes
\begin{equation}
L=L_{\mathrm{Rec}}+w_{\mathrm{Acc}}L_{\mathrm{Acc}},
\end{equation}
where $w_{\mathrm{Acc}}$ weights Accelerated Distillation.

\paragraph{Theoretical analysis.}
To understand how the two local distillation terms recover longer refinement chains, we model sequence-level distillation with exact distributional backups, following the one-step bootstrap structure of temporal-difference learning \citep{bellman1957dp,sutton1988td}.

\begin{theorem}[Compression and amplification]
\label{thm:main}
After $n$ exact distributional backups of Recursive Distillation, the model's $k$-th attempt matches the initial model's $(n{+}1)k$-th attempt in distribution.
Answer distillation compresses later attempts into the first, while refinement distillation makes each refinement step equivalent to $n{+}1$ initial-model steps.
\end{theorem}

Accelerated Distillation lets the first attempt use this strengthened refinement operation directly, accumulating its gains across updates.

\begin{proposition}[Accelerated Distillation]
\label{prop:accelerated}
Accelerated Distillation makes the first attempt a mixture of initial-model attempts whose expected index grows quadratically with the number of exact backups.
\end{proposition}

Together, these results show that local supervision strengthens refinement and moves the first attempt toward the initial model's infinite-step refinement limit, with Accelerated Distillation speeding the transfer (proofs in \cref{app:theory}).

\icrsi and \iwrsi thus close the loop: teacher guidance strengthens answering and refinement, and the updated policy generates the next reference solutions.
This turns test-time compute into lasting model improvement without external labels.

\section{Experiments}
\label{sec:experiments}

We compare \ttrsi with existing label-free test-time training methods (\cref{sec:exp-main}), then ablate its components (\cref{sec:exp-ablation}) and analyze its gains and costs (\cref{sec:exp-analysis}).

\begin{table*}[t]
\centering
\small
\microtypesetup{protrusion=false}
\setlength{\tabcolsep}{4pt}
\renewcommand{\arraystretch}{0.94}
\setlength{\aboverulesep}{1.5pt}
\setlength{\belowrulesep}{2pt}
\begin{tabularx}{\textwidth}{l*{5}{>{\centering\arraybackslash}X}c}
\toprule
Method & AIME25 & HMMT25 & AIME26 & HMMT26 & BRUMO25 & Avg. \\
\midrule
\multicolumn{7}{l}{\textbf{Qwen3-1.7B}} \\
Base & 37.78 & 21.67 & 39.17 & 27.27 & 48.33 & 34.84 \\
TTRL \citep{zuo2025ttrl} & 36.39 & 25.56 & 41.67 & 28.54 & 48.61 & 36.15 \\
TTPO \citep{wang2026ttpo} & \underline{41.94} & 27.22 & \underline{46.67} & 31.82 & \underline{52.22} & \underline{39.97} \\
OPSD-TTT \citep{wang2026ttpo} & 40.83 & \underline{28.89} & 45.28 & \underline{32.58} & 51.67 & 39.85 \\
CANON \citep{gkountouras2026canon} & 33.61 & 21.94 & 31.67 & 27.78 & 49.72 & 32.94 \\
U-OPSD \citep{uopsd2026} & 40.00 & 25.00 & 43.06 & 29.80 & 48.33 & 37.24 \\
\ourmethodrow
\iwrsi (ours) & \textbf{44.72} & \textbf{29.17} & \textbf{48.61} & \textbf{33.84} & \textbf{53.61} & \textbf{41.99}\nobreak\hspace{0.3em}\scoregain{+7.15} \\
\arrayrulecolor{black!35}\cmidrule(lr){1-7}\arrayrulecolor{black}
Base + \icrsi & 36.67 & \underline{30.00} & 50.00 & 30.30 & 51.67 & 39.73 \\
TTPO \citep{wang2026ttpo} + SC & \underline{43.33} & \underline{30.00} & \underline{56.67} & \underline{33.33} & \textbf{60.00} & \underline{44.67} \\
\ourmethodrow
\ttrsi (ours) & \textbf{53.33} & \textbf{36.67} & \textbf{63.33} & \textbf{42.42} & \underline{56.67} & \textbf{50.48}\nobreak\hspace{0.3em}\scoregain{+15.64} \\
\midrule
\multicolumn{7}{l}{\textbf{Qwen3-4B}} \\
Base & 68.06 & 40.28 & 65.00 & 42.68 & 62.78 & 55.76 \\
TTRL \citep{zuo2025ttrl} & 68.61 & 44.72 & 66.11 & \underline{44.95} & 65.83 & 58.04 \\
TTPO \citep{wang2026ttpo} & \underline{71.39} & \underline{47.78} & \underline{68.89} & 43.18 & 66.67 & \underline{59.58} \\
OPSD-TTT \citep{wang2026ttpo} & 70.28 & 46.39 & \underline{68.89} & 42.93 & \underline{68.33} & 59.36 \\
CANON \citep{gkountouras2026canon} & 62.50 & 40.56 & 63.33 & 37.63 & 65.00 & 53.80 \\
U-OPSD \citep{uopsd2026} & 67.78 & 44.17 & 66.94 & 41.67 & 64.72 & 57.06 \\
\ourmethodrow
\iwrsi (ours) & \textbf{72.50} & \textbf{48.33} & \textbf{72.50} & \textbf{46.72} & \textbf{69.44} & \textbf{61.90}\nobreak\hspace{0.3em}\scoregain{+6.14} \\
\arrayrulecolor{black!35}\cmidrule(lr){1-7}\arrayrulecolor{black}
Base + \icrsi & 76.67 & \underline{47.78} & \underline{76.67} & \underline{50.51} & 70.00 & 64.33 \\
TTPO \citep{wang2026ttpo} + SC & \underline{80.00} & \textbf{53.33} & \underline{76.67} & 45.45 & \underline{73.33} & \underline{65.76} \\
\ourmethodrow
\ttrsi (ours) & \textbf{83.33} & \textbf{53.33} & \textbf{83.33} & \textbf{51.52} & \textbf{76.67} & \textbf{69.64}\nobreak\hspace{0.3em}\scoregain{+13.88} \\
\midrule
\multicolumn{7}{l}{\textbf{Qwen3-8B}} \\
Base & 70.56 & 46.39 & 67.78 & 45.20 & 70.56 & 60.10 \\
TTRL \citep{zuo2025ttrl} & 68.61 & 46.67 & \underline{71.11} & 45.96 & \textbf{73.89} & 61.25 \\
TTPO \citep{wang2026ttpo} & \underline{72.22} & 47.22 & 70.83 & 47.22 & 71.67 & \underline{61.83} \\
OPSD-TTT \citep{wang2026ttpo} & 70.83 & \underline{47.78} & 70.28 & \underline{47.73} & 71.11 & 61.55 \\
CANON \citep{gkountouras2026canon} & 66.39 & 43.89 & 67.78 & 44.95 & 70.28 & 58.66 \\
U-OPSD \citep{uopsd2026} & 69.72 & \textbf{49.17} & 70.56 & 47.22 & 70.56 & 61.45 \\
\ourmethodrow
\iwrsi (ours) & \textbf{74.44} & \textbf{49.17} & \textbf{74.44} & \textbf{49.49} & \underline{73.33} & \textbf{64.17}\nobreak\hspace{0.3em}\scoregain{+4.07} \\
\arrayrulecolor{black!35}\cmidrule(lr){1-7}\arrayrulecolor{black}
Base + \icrsi & 76.67 & \underline{50.00} & \underline{76.67} & \underline{51.52} & \underline{76.67} & 66.31 \\
TTPO \citep{wang2026ttpo} + SC & \textbf{83.33} & \underline{50.00} & 73.33 & \textbf{54.55} & \underline{76.67} & \underline{67.58} \\
\ourmethodrow
\ttrsi (ours) & \underline{80.00} & \textbf{56.67} & \textbf{86.67} & \textbf{54.55} & \textbf{83.33} & \textbf{72.24}\nobreak\hspace{0.3em}\scoregain{+12.14} \\
\bottomrule
\end{tabularx}
\caption{\textbf{Main results.}
We report accuracy (\%) across three model scales and five mathematics benchmarks. For test-time scaling, \ttrsi applies \icrsi to the \iwrsi model and reports the final attempt, while TTPO + SC applies Self-Consistency \citep{wang2023selfconsistency} to the TTPO model with the same budget.
Avg.\ averages the benchmarks, and blue parentheses give accuracy gains over Base.
Performances are illustrated with \textbf{best} and \underline{second best} within each group, including ties.}
\label{tab:main}
\end{table*}

\subsection{Setup}
\label{sec:exp-setup}

\paragraph{Models and benchmarks.}
We evaluate Qwen3-1.7B, 4B and 8B \citep{yang2025qwen3} on five competition mathematics benchmarks: AIME~2025 and 2026, HMMT February~2025 and 2026, and BRUMO~2025 \citep{balunovic2025matharena,dekoninck2026matharena}.
Following prior label-free test-time training \citep{zuo2025ttrl,wang2026ttpo}, our main comparison adapts each method separately to the unlabeled questions of each benchmark.
Labels are used only for evaluation.
We report first- and final-attempt accuracy as avg@12.
Responses are sampled with temperature 1.0 and top-$p$ 0.95, generating up to 38{,}912 tokens (\cref{app:eval}).

\paragraph{Baselines.}
We compare with TTRL \citep{zuo2025ttrl}, TTPO and OPSD-TTT \citep{wang2026ttpo}, CANON \citep{gkountouras2026canon}, and U-OPSD \citep{uopsd2026}, covering majority-based policy optimization and label-free self-distillation (\cref{app:baselines}).
We also compare with TTPO using self-consistency (TTPO + SC) under the same test-time scaling budget as \icrsi (\cref{app:budget}).

\paragraph{Implementation.}
In our main experiments, \icrsi uses chains of $\ell{=}11$ attempts, that is, one initial solution and ten refinements.
\iwrsi uses 100 training updates with 32 question occurrences per update.
All three loss terms have unit weight ($w_{\mathrm{Acc}}{=}1$) and use the OPSD forward KL with temperature 1.1, clipping the per-token KL at 0.05.
We use $\alpha{=}0.5$ and $\tau{=}10^{-4}$.
We design a pipelined sampling schedule that overlaps generation across updates, reducing idle time and accelerating training (\cref{app:pipeline}).

\subsection{Main results}
\label{sec:exp-main}

First, under the same test-time scaling budget, \ttrsi outperforms self-consistency-based test-time training at every model scale (see \cref{tab:main}).
On average, it exceeds TTPO + SC by \textbf{+4.79\%}, with wins in 11 of the 15 settings and ties in two.
On AIME26 at 8B, for example, accuracy is \textbf{86.67\%}, compared with 73.33\% for TTPO + SC (\textbf{+13.34\%}).
Self-consistency-based training can reinforce an incorrect majority, so further sampling tends to reproduce the same error: in the same setting, self-consistency lifts TTPO by only \textbf{+2.50\%}, whereas \icrsi lifts Base by \textbf{+8.89\%}.
Training on refinements instead strengthens revision, improving the solutions that later attempts build on.

Second, even without test-time scaling, \iwrsi achieves the best first-attempt accuracy among label-free training methods at every model scale.
It improves over Base by \textbf{+5.79\%} on average, beating all five training baselines in 13 of the 15 settings and tying for the best score in one other.
At 8B, its mean accuracy is \textbf{64.17\%}, \textbf{+2.34\%} above TTPO \citep{wang2026ttpo}, the strongest training baseline on average.
These gains show that distillation retains refinement benefits in the model and remains effective with a stronger starting policy.

Third, the gains from training also carry over to inference-time refinement.
Specifically, the full framework gains \textbf{+13.89\%} over Base and exceeds Base + \icrsi by \textbf{+7.33\%}.
\iwrsi therefore helps even when in-context improvement is available at inference, raising both the first and later attempts.
Notably, at 1.7B, the first attempt of \iwrsi (41.99\%) already exceeds Base + \icrsi after ten refinements (39.73\%), consistent with the compression of later attempts into the first attempt (\cref{thm:main}).

\subsection{Ablations}
\label{sec:exp-ablation}

We consider each component in turn, first the refinement design of \icrsi and then the distillation design of \iwrsi.

\begin{table}[t]
\centering
\small
\setlength{\tabcolsep}{3pt}
\begin{adjustbox}{max width=\linewidth}
\begin{tabular}{lcccc}
\toprule
Setting & Acc. & Fix & Break & $\overline{f/(f{+}b)}$ \\
\midrule
Base & 53.84 & -- & -- & -- \\
Final answer only & 59.31 & 7.01 & 3.95 & 60.55 \\
Directed reference use & 60.43 & 2.12 & 1.24 & 59.97 \\
Unrelated reference & 50.61 & 14.43 & 13.80 & 49.57 \\
\ourmethodrow
\icrsi (ours) & \textbf{63.59} & 3.84 & 1.96 & \textbf{65.42} \\
\bottomrule
\end{tabular}
\end{adjustbox}
\caption{\textbf{\icrsi design ablations.}
\mbox{Qwen3-4B} accuracy, fix/break rates and balance points (\%), averaged over AIME26 and HMMT26.
Balance points are computed per benchmark before averaging.
The \textbf{best} accuracy and balance point are in bold.}
\label{tab:ladder}
\end{table}

\paragraph{\icrsi design.}
To examine how the two design choices of \icrsi (see \cref{sec:icrsi}) affect refinement, we compare it with three variants (\cref{tab:ladder}).
\emph{Final answer only} retains just the final answer of $a_k$, \emph{directed reference use} instructs the policy to check and correct $a_k$, and \emph{unrelated reference} replaces $a_k$ with a solution to a different question from the same benchmark.
Alongside accuracy, we report the fix rates, break rates and balance points of \cref{prop:dyn}, with rates pooled over the chain and estimated in a separate evaluation batch.
The full design raises mean accuracy from \textbf{53.84\%} to \textbf{63.59\%} after ten refinement steps.
Replacing the full solution with only its final answer lowers mean accuracy (\textbf{\textminus4.28\%}) and roughly doubles the break rate, indicating that intermediate reasoning helps preserve correct solutions.
Directing reference use lowers accuracy (\textbf{\textminus3.16\%}) and the fix rate, suggesting that autonomous use leaves more room to correct earlier mistakes.
An~unrelated reference performs below Base (\textbf{\textminus3.23\%}), showing that additional context alone does not provide the benefit of relevant prior reasoning.
The two design choices together improve refinement.

The balance points reflect the same pattern.
Unrelated references have the highest fix rate but also introduce frequent regressions, yielding the lowest balance point.
Full-solution references and autonomous use balance correction with preservation, giving \icrsi the highest balance point.
Across the four settings, these balance points track observed accuracy within 1.83\%.

\begin{table}[t]
\centering
\small
\renewcommand{\arraystretch}{1.28}
\setlength{\tabcolsep}{3pt}
\begin{adjustbox}{max width=\linewidth}
\begin{tabular}{lcc}
\toprule
Variant & AIME26 & HMMT26 \\
\midrule
\multicolumn{3}{l}{\emph{(a) Distillation objective}} \\
Answer distillation ($L_{\mathrm{ans}}$) only & 68.61 & 44.19 \\
Refinement distillation ($L_{\mathrm{ref}}$) only & 69.44 & 44.95 \\
Recursive Distillation & 71.39 & \underline{45.96} \\
\ourmethodrow
\quad \textbf{+ Accelerated Distillation} (ours) & \underline{72.50} & \textbf{46.72} \\
\iwrsi + third pair $(a_2,a_3)$ & \textbf{72.78} & \textbf{46.72} \\
\midrule
\multicolumn{3}{l}{\emph{(b) Reference policy ($a_1\to a_{11}$)}} \\
Fixed initial & $\underline{70.83}\to\underline{74.44}$ & $\underline{42.93}\to\underline{47.73}$ \\
\ourmethodrow
Current (ours) & $\mathbf{72.50}\to\mathbf{83.33}$ & $\mathbf{46.72}\to\mathbf{51.52}$ \\
\bottomrule
\end{tabular}
\end{adjustbox}
\caption{\textbf{\iwrsi design ablations.}
Qwen3-4B accuracy (\%) for different distillation objectives (a) and reference policies (b).
Performances are illustrated with \textbf{best} and \underline{second best}.}
\label{tab:ablation}
\label{tab:reference-policy}
\end{table}

\paragraph{\iwrsi design.}
To test whether supervising both behaviors (\cref{sec:iwrsi}) improves on either alone, we train the two distillation terms separately and together (see \cref{tab:ablation}a).
Recursive Distillation outperforms refinement distillation, the stronger single term, by \textbf{+1.95\%} on AIME26 and \textbf{+1.01\%} on HMMT26.
We then add Accelerated Distillation to test whether coupling improves the first attempt, obtaining further gains of \textbf{+1.11\%} and \textbf{+0.76\%}, respectively.
Training both behaviors and coupling their guidance therefore strengthen the first attempt.
Finally, to assess the value of deeper supervision, we add a third adjacent pair, in which a teacher given $a_3$ supervises a student given $a_2$.
It yields only \textbf{+0.28\%} on AIME26 and matches \iwrsi on HMMT26 (\cref{prop:deeper} in \cref{app:theory}), while requiring another generation.
We therefore retain two reference solutions per update.

\subsection{Analysis}
\label{sec:exp-analysis}

Having shown that training strengthens refinement, we now examine how these gains feed back through references, how far they extend, whether they reach questions that the majority gets wrong, what they cost, and how they accumulate during adaptation.

\paragraph{Mutual reinforcement across updates.}
We compare references generated by the current and initial policies under the same distillation objective (see \cref{tab:reference-policy}b).
Generating references with the evolving policy improves first-attempt accuracy by \textbf{+1.67\%} on AIME26 and \textbf{+3.79\%} on HMMT26 over using the fixed initial policy.
After ten refinement steps, the advantages are \textbf{+8.89\%} and \textbf{+3.79\%}, respectively.
Evolving references therefore improve both the first and final attempts.
This matches the bootstrap analysis (\cref{thm:main}): fixed initial-policy references always provide the same one-step targets, whereas evolving references let the gains compound across updates.
Together with the stronger refinement observed after training (\cref{sec:exp-main}), this shows how improvements in the policy and its references reinforce one another.

\begin{table}[t]
\centering
\small
\renewcommand{\arraystretch}{1.034}%
\setlength{\tabcolsep}{4pt}
\begin{adjustbox}{max width=\linewidth}
\begin{tabular}{lcccc}
\toprule
\multicolumn{5}{l}{\textbf{Train on AIME26}} \\[0.5pt]
Method & HMMT25 & HMMT26 & BRUMO25 & Avg. \\
\midrule
Base & 40.28 & 42.68 & 62.78 & 48.58 \\
TTRL \citep{zuo2025ttrl} & 40.83 & 40.15 & 62.22 & 47.73 \\
TTPO \citep{wang2026ttpo} & \underline{43.33} & 42.68 & 64.44 & 50.15 \\
OPSD-TTT \citep{wang2026ttpo} & \underline{43.33} & \underline{42.93} & \underline{65.28} & \underline{50.51} \\
CANON \citep{gkountouras2026canon} & 38.61 & 39.65 & 64.72 & 47.66 \\
U-OPSD \citep{uopsd2026} & 42.22 & 40.91 & 64.17 & 49.10 \\
\ourmethodrow
\iwrsi (ours) & \textbf{43.61} & \textbf{43.69} & \textbf{65.56} & \textbf{50.95} \\
\midrule
\multicolumn{5}{l}{\textbf{Train on HMMT26}} \\[0.5pt]
Method & AIME25 & AIME26 & BRUMO25 & Avg. \\
\midrule
Base & 68.06 & 65.00 & 62.78 & 65.28 \\
TTRL \citep{zuo2025ttrl} & 68.33 & 63.06 & 63.61 & 65.00 \\
TTPO \citep{wang2026ttpo} & \underline{69.17} & 66.67 & \underline{66.67} & 67.50 \\
OPSD-TTT \citep{wang2026ttpo} & \underline{69.17} & \underline{68.06} & 65.83 & \underline{67.69} \\
CANON \citep{gkountouras2026canon} & 60.83 & 61.11 & 59.17 & 60.37 \\
U-OPSD \citep{uopsd2026} & 66.67 & 66.11 & 65.83 & 66.20 \\
\ourmethodrow
\iwrsi (ours) & \textbf{69.44} & \textbf{68.33} & \textbf{67.22} & \textbf{68.33} \\
\bottomrule
\end{tabular}
\end{adjustbox}
\caption{\textbf{Cross-task transfer.}
Qwen3-4B accuracy (\%) on target benchmarks after adaptation to AIME26 or HMMT26.
Performances are illustrated with \textbf{best} and \underline{second best}.}
\label{tab:transfer}
\end{table}

\paragraph{Cross-task transfer.}
To test whether improvements from training extend beyond the adaptation questions, we train Qwen3-4B on AIME26 or HMMT26 and evaluate on the target benchmarks without further updates (\cref{tab:transfer}).
Compared with Base, mean target accuracy improves by \textbf{+2.37\%} and \textbf{+3.05\%}, respectively.
\iwrsi achieves the highest accuracy on all six source--target pairs, averaging \textbf{59.64\%} and exceeding OPSD-TTT, the strongest baseline on average.
These transfer gains show that the improvement is retained in the model weights and applies to new questions.

\begin{table}[t]
\centering
\small
\renewcommand{\arraystretch}{0.965}%
\setlength{\tabcolsep}{3pt}
\begin{adjustbox}{max width=\linewidth}
\begin{tabular}{lcccc}
\toprule
\multicolumn{5}{l}{\emph{(a) Beyond the initial majority (AIME26)}} \\
& \multicolumn{2}{c}{All questions} & \multicolumn{2}{c}{Wrong majority} \\
\cmidrule(lr){2-3}\cmidrule(lr){4-5}
Method & avg@64 & SC@64 & avg@64 & SC correct \\
\midrule
Base & 66.56 & \textbf{80.00} & 5.73 & 0/6 \\
TTRL \citep{zuo2025ttrl} & 64.48 & \underline{76.67} & 4.95 & 0/6 \\
TTPO \citep{wang2026ttpo} & 67.34 & \underline{76.67} & 5.47 & 0/6 \\
OPSD-TTT \citep{wang2026ttpo} & \underline{67.66} & \underline{76.67} & \underline{8.33} & \underline{1/6} \\
CANON \citep{gkountouras2026canon} & 61.25 & 70.00 & 2.34 & 0/6 \\
U-OPSD \citep{uopsd2026} & 64.79 & \underline{76.67} & 2.08 & 0/6 \\
\ourmethodrow
\iwrsi (ours) & \textbf{71.35} & \textbf{80.00} & \textbf{12.76} & \textbf{2/6} \\
\midrule
\multicolumn{5}{l}{\emph{(b) Training efficiency}} \\
& \multicolumn{2}{c}{AIME26} & \multicolumn{2}{c}{HMMT26} \\
\cmidrule(lr){2-3}\cmidrule(lr){4-5}
Method & Time$\downarrow$ & Tokens$\downarrow$ & Time$\downarrow$ & Tokens$\downarrow$ \\
\midrule
TTRL \citep{zuo2025ttrl} & \underline{22.08} & \underline{1.63} & \underline{23.08} & \underline{1.74} \\
TTPO \citep{wang2026ttpo} & 45.73 & 11.74 & 37.70 & 10.89 \\
U-OPSD \citep{uopsd2026} & 37.27 & 4.59 & 50.30 & 5.56 \\
\ourmethodrow
\iwrsi (ours) & \textbf{8.53} & \textbf{0.95} & \textbf{10.06} & \textbf{1.12} \\
\bottomrule
\end{tabular}
\end{adjustbox}
\caption{\textbf{Initial-majority behavior and training cost.}
Qwen3-4B accuracy (\%) and correct-majority counts (a), and minutes and tokens (M) per update (b).
Performances are illustrated with \textbf{best} and \underline{second best}.}
\label{tab:axes}
\label{tab:training-efficiency}
\end{table}

\paragraph{Gains beyond the initial majority.}
To test whether refinement-based training can overcome an initially wrong majority, we evaluate the \iwrsi model and the baselines with 64 independent answers per question on AIME26, focusing on the six questions whose Base majority is wrong (\cref{tab:axes}a), where avg@64 averages the 64 answers and SC@64 is their majority vote.
Several self-consistency-based baselines perform below Base on this subset, illustrating how training on agreement can reinforce an initially wrong answer.
On these questions, \iwrsi raises avg@64 from 5.73\% to \textbf{12.76\%} (\textbf{+7.03\%}), exceeding OPSD-TTT's 8.33\% by \textbf{+4.43\%}.
Self-consistency also becomes correct on two of these questions, compared with one for OPSD-TTT.
Overall avg@64 improves by \textbf{+4.79\%} over Base and \textbf{+3.69\%} over OPSD-TTT, the strongest baseline in this comparison.
This demonstrates that training on refinements raises accuracy on questions whose majority is wrong, not only on those the model already answers correctly by majority.

\begin{figure}[t]
\centering
\includegraphics[width=\linewidth]{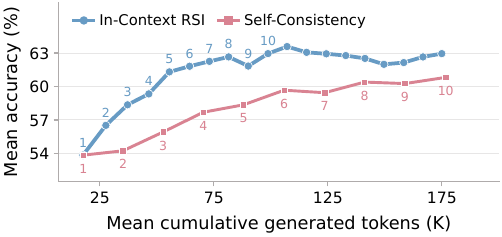}
\caption{\textbf{\icrsi vs.\ Self-Consistency.}
Mean accuracy of \mbox{Qwen3-4B} versus mean cumulative generated tokens over AIME26 and HMMT26.}
\label{fig:tts}
\end{figure}

\paragraph{Inference efficiency.}
To compare \icrsi with Self-Consistency at inference, we plot mean accuracy over AIME26 and HMMT26 against mean cumulative generated tokens per question (\cref{fig:tts}).
\icrsi exceeds the accuracy of ten-sample self-consistency at its fifth attempt while generating \textbf{68.6\%} fewer tokens, demonstrating more efficient inference scaling without weight updates.
At comparable total token cost, it retains an accuracy advantage of \textbf{+2.13\%}.

\paragraph{Training efficiency.}
To measure the training cost of the improvement loop, we compare methods with 32 question occurrences per update on four H20 GPUs (\cref{tab:training-efficiency}b).
With pipelined reference generation, \iwrsi takes \textbf{8.53} minutes on AIME26 and \textbf{10.06} on HMMT26, giving a speedup of \textbf{2.29}--\textbf{5.36$\times$} per update over the baselines.
It also uses the fewest tokens per update, \textbf{0.95M} on AIME26 and \textbf{1.12M} on HMMT26, reducing the cost of its self-supervision.
Together with the inference savings above, these results show reduced training time and inference token cost.

\begin{figure}[t]
\centering
\includegraphics[width=\linewidth]{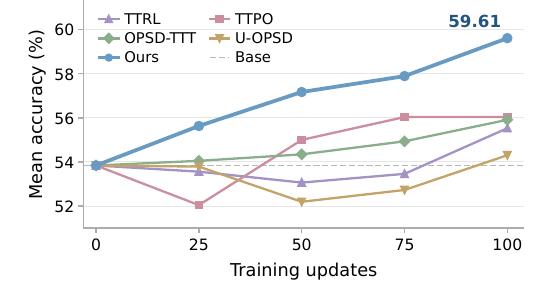}
\caption{\textbf{Performance during adaptation.}
We track Qwen3-4B mean accuracy over AIME26 and HMMT26, with the dashed line marking Base.}
\label{fig:training-dynamics}
\end{figure}

\paragraph{Performance during adaptation.}
Beyond per-update cost, we track first-attempt accuracy across training updates to see how refinement benefits accumulate in the model's weights (\cref{fig:training-dynamics}).
\iwrsi leads throughout training, reaching \textbf{59.61\%} accuracy (\textbf{+5.77\%} over Base).
Training on successive refinements therefore continues to improve the model throughout adaptation.

\begin{figure}[t]
\centering
\input{figures/case_studies}
\caption{\textbf{Training accelerates refinement.}
Self-consistency selects a wrong most-frequent answer on an HMMT26 problem, while refinement succeeds at $a_{11}$ before training and $a_2$ after \iwrsi training.
Blue marks reasoning revisions.}
\label{fig:case-studies}
\end{figure}

\paragraph{Case study.}
Finally, to inspect how refinement changes the reasoning, we trace the stick-breaking example in \cref{fig:case-studies} (\cref{app:case-studies}).
Self-consistency fails because its most frequent answer, produced by three of eleven samples, comes from a flawed averaging argument and outvotes the two samples reaching the correct $63/256$, while base refinement succeeds by replacing that argument with a recurrence for $S(n)$, the expected sum of squared piece lengths.
After \iwrsi training, the chain corrects its transition factor at $a_2$ using the number of pieces and remains correct thereafter.
Reaching a correct derivation at $\boldsymbol{a_2}$ rather than $\boldsymbol{a_{11}}$ illustrates the amplification of refinement described in our analysis (\cref{thm:main}).

\section{Related work}
\label{sec:related}

\paragraph{Test-time scaling.}
Parallel scaling aggregates independent solutions \citep{wang2023selfconsistency,brown2024monkeys}, while sequential methods build on earlier attempts by revision or re-solving \citep{madaan2023selfrefine,tian2025thinktwice}.
Repetition and limited exploration can restrict these gains \citep{parseq2026gap,seqexplore2026}.
Compute can also be allocated within a single attempt, by its reasoning length \citep{guo2025deepseekr1} or by the effective depth of the architecture \citep{liang2026asymloop}, rather than by the number of attempts.
\icrsi preserves exploration by letting the model decide whether and how to use full-solution references, and \iwrsi distills the resulting refinements.

\paragraph{Label-free test-time training.}
This line of work adapts models on unlabeled test inputs using self-generated supervision \citep{sun2020ttt,akyurek2025ttt}.
TTRL rewards majority agreement \citep{zuo2025ttrl}, and TTPO distills solutions selected by self-consistency \citep{wang2026ttpo}.
Other methods use agreement to select teacher references \citep{gkountouras2026canon,uopsd2026} or sharpen confidence \citep{zhao2026intuitor,agarwal2025entropy}.
\iwrsi uses successive solutions as teacher references, obtaining guidance from refinement rather than majority agreement.

\paragraph{Self-improvement and distillation.}
Training on self-generated attempts has been studied with correctness feedback \citep{zelikman2022star,kumar2025score} and in recursive training frameworks \citep{mass2026}, which a recent survey organizes by how far the improvement loop is closed \citep{rsibounded2026}.
Recent systems evolve algorithms and agents \citep{novikov2025alphaevolve,zhang2026dgm} or consolidate deployment experience into model parameters \citep{ye2026oel}.
On-policy self-distillation uses contextual information to construct dense teacher supervision \citep{zhao2026opsd,wu2026dapd}.
\ttrsi connects in-context improvement with weight updates by distilling successive self-generated refinements, forming a test-time amplify-then-distill loop \citep{anthony2017exit,christiano2018ida}.

\section{Conclusion}
\label{sec:conclusion}

We presented \ttrsi, a label-free framework for improving LLMs at test time.
\icrsi lets the model autonomously build on prior solutions, which \iwrsi uses as teacher references to strengthen answering and refinement.
Our analysis shows how local distillation compresses and amplifies improvement toward the model's infinite-step refinement limit.
Across three model scales and five benchmarks, \ttrsi gains \textbf{+13.89\%} over Base and \textbf{+4.79\%} over TTPO + SC on average.

\section*{Limitations}

Our evaluation covers mathematical reasoning and the Qwen3 family up to 8B parameters.
A broader direction is to extend \ttrsi to open-ended tasks and tool-using agents, where improvement unfolds over interactions and must be retained across tasks.

\clearpage
\bibliography{references}

\clearpage
\appendix
\crefalias{section}{appendix}\crefalias{subsection}{appendix}
\section{Theory: details and proofs}
\label{app:theory}

\subsection{Solution distributions and the one-step backup}
As in the main text, the teacher uses the frozen initial parameters $\theta_0$.
Fix a question $q$.
Let $\A$ be the finite set of possible solutions (token sequences up to the generation limit).
We write distributions as row vectors and kernels as row-stochastic matrices.
Thus $\nu K$ is the distribution obtained by applying $K$ to a sample from $\nu$, and $\one\nu$ is the kernel whose every row is $\nu$.
We use $\TV{\nu-\nu'}=\frac12\sum_a|\nu(a)-\nu'(a)|$ and define $\acc(\nu)$ as the total probability of solutions with a correct final answer.
We write $\mu_\theta(a)=\pi_\theta(a\mid q)$ for the first-attempt distribution and $K_\theta(a,a')=\pi_\theta(a'\mid q,a)$ for the improvement kernel.
The $k$-th attempt has distribution $\mu_\theta K_\theta^{k-1}$.
Let $\mu_0=\mu_{\theta_0}$ and $T=K_{\theta_0}$ denote the initial model's first-attempt distribution and improvement kernel.
Its $k$-th attempt has distribution $\mu_0T^{k-1}$, and $T$ is the fixed teacher operator in \cref{eq:adjacent,eq:accelerated}.
The Dobrushin coefficient of a kernel, $\dob(K)=\max_{a,a'}\TV{K(a,\cdot)-K(a',\cdot)}$, satisfies $\TV{\nu K-\nu'K}\le\dob(K)\TV{\nu-\nu'}$ and $\dob(KK')\le\dob(K)\dob(K')$ \citep{dobrushin1956,levin2017markov}.
We call $T$ \emph{ergodic} if $\dob(T^m)<1$ for some $m\ge1$.
It then has a unique stationary distribution $\pinf$, the infinite-step limit of the initial model's refinement chain.
Indeed, the map $\nu\mapsto\nu T^m$ is a contraction of the probability simplex, so it has a unique fixed point.
Applying $T$ to that fixed point gives another fixed point of $T^m$, hence the same distribution, proving stationarity for $T$.
Writing $s=jm+r$ gives convergence of $\nu T^s$ to $\pinf$ from every $\nu$.
This mixing condition is used for convergence. The finite-update identities hold for every stochastic $T$.

For any solution distribution $\nu$, the one-step backup is
\begin{equation}
\mathcal{B}_T\nu=\nu T=\E_{a\sim\nu}T(a,\cdot).
\end{equation}
Recursive Distillation applies this backup to $\mu_\theta$ and to each row of $K_\theta$.

\subsection{The update}
The full objective is
\begin{align}
L_{\mathrm{Rec}}&=L_{\mathrm{ans}}+L_{\mathrm{ref}},\nonumber\\
L&=L_{\mathrm{Rec}}+w_{\mathrm{Acc}}L_{\mathrm{Acc}}.
\end{align}
The coefficient $w_{\mathrm{Acc}}\ge0$ controls Accelerated Distillation, with $w_{\mathrm{Acc}}=0$ giving Recursive Distillation alone.

\begin{lemma}[Mixture minimizer]
\label{lem:mixture}
For distributions $\nu_1,\dots,\nu_k$ and weights $w_i\ge0$ with $W=\sum_iw_i>0$, $\sum_iw_i\KL(\nu_i\|\nu)$ is uniquely minimized by $\bar\nu=\frac1W\sum_iw_i\nu_i$.
\end{lemma}
\begin{proof}
$\sum_iw_i\KL(\nu_i\|\nu)=\sum_iw_i\KL(\nu_i\|\bar\nu)+W\KL(\bar\nu\|\nu)$.
By Gibbs' inequality, this is uniquely minimized at $\nu=\bar\nu$.
\end{proof}

We use a common solution-distribution model for references and student rollouts, abstracting their different generation modes.
To derive the backup, we replace the token-level loss by sequence-level forward KL and hold the reference-sampling distributions fixed during each update.
With $a_1\sim\mu_\theta$ and $a_2\sim K_\theta(a_1,\cdot)$, the component losses in \cref{eq:adjacent,eq:accelerated} become as follows.
Answer distillation is $L_{\mathrm{ans}}(\mu)=\E\,\KL(T(a_1,\cdot)\|\mu)$, and Accelerated Distillation is $L_{\mathrm{Acc}}(\mu)=\E\,\KL(T(a_2,\cdot)\|\mu)$.
Refinement distillation is $L_{\mathrm{ref}}(K)=\E\,\KL(T(a_2,\cdot)\|K(a_1,\cdot))$.
Let $\gamma=w_{\mathrm{Acc}}/(1+w_{\mathrm{Acc}})\in[0,1)$ be the constant share of the accelerated target in the first-attempt backup, with $\gamma=1/2$ when $w_{\mathrm{Acc}}=1$ and $\gamma=0$ for Recursive Distillation alone.
By \cref{lem:mixture}, the minimizer of $(1-\gamma)L_{\mathrm{ans}}+\gamma L_{\mathrm{Acc}}$ is $(1-\gamma)\mu_\theta T+\gamma\mu_\theta K_\theta T$.
$L_{\mathrm{ref}}$ decomposes over rows: every row $a$ with positive sampling probability has minimizer $(K_\theta T)(a,\cdot)$.
The following full backup applies this target to each row, treating solution distributions as the optimization variables.

\begin{definition}[Exact distributional backup]
\label{def:backup}
For $0\le\gamma<1$, one full backup is
\begin{equation}
\Phi(\mu,K)=\big(\mu((1-\gamma)I+\gamma K)T,\;KT\big).
\label{eq:Phi}
\end{equation}
Without Accelerated Distillation, $\Phi(\mu,K)=(\mu T,KT)$.
\end{definition}

We write $(\mu_n,K_n)=\Phi^n(\mu_0,T)$ for the distributions after $n$ exact backups.

\paragraph{One-step bootstrapping.}
For $A\sim\nu$, a single reference provides an unbiased target, $\E[T(A,\cdot)\mid\nu]=\nu T$.
Taking $\nu=\mu$ or $\nu=K(a,\cdot)$ gives the first-attempt or refinement backup, respectively.
This is the local one-step bootstrap structure used in temporal-difference learning \citep{sutton1988td}.

The analysis uses sequence-level forward KL to characterize the bootstrap dynamics.
Training implements distillation at student-generated prefixes, with token weighting and clipping.

\subsection{Proofs of the main results}
\begin{proof}[Proof of \cref{prop:dyn}]
Let $\acc_k=\acc(\mu K^{k-1})$.
By the definitions of the fix and break rates, $\acc_{k+1}=(1-\acc_k)f+\acc_k(1-b)=f+(1-f-b)\,\acc_k$.
With $\acc_\infty=f/(f{+}b)$, this gives $\acc_{k+1}-\acc_\infty=(1-f-b)(\acc_k-\acc_\infty)$, and iterating yields
\begin{equation}
\acc_k=\acc_\infty+(1{-}f{-}b)^{k-1}(\acc_1-\acc_\infty).
\end{equation}
Because $1-f-b\in[0,1)$, $\acc_k$ moves monotonically toward $\acc_\infty$, so $\acc_k>\acc_1$ for $k\ge2$ if and only if $\acc_\infty>\acc_1$.
\end{proof}

We prove \cref{thm:main} and \cref{prop:accelerated} together, since \cref{thm:main} is the case $\gamma=0$.
For Accelerated Distillation, the expected equivalent initial-model attempt index is $1+n+\gamma\,n(n{+}1)/2$.

\begin{proof}[Proof of \cref{thm:main} and \cref{prop:accelerated}]\leavevmode\par
\emph{Closed form.}
The update of $K$ in \cref{eq:Phi} does not involve $\mu$, so $K_n=T^{n+1}$ by induction from $K_0=T$.
Substituting into the update of $\mu$ gives $\mu_{i+1}=\mu_i((1-\gamma)T+\gamma T^{i+2})$, hence
\begin{equation}
\mu_n=\mu_0\prod_{i=0}^{n-1}((1-\gamma)T+\gamma T^{i+2}).
\label{eq:accelerated-closed}
\end{equation}
Each factor is a mixture of $T$ (probability $(1-\gamma)$) and $T^{i+2}$ (probability $\gamma$).
Thus $\mu_n=\E[\mu_0T^{S}]$, where $S=\sum_{i<n}s_i$, the $s_i$ are independent with $\Pr(s_i=1)=(1-\gamma)$ and $\Pr(s_i=i+2)=\gamma$, and $\E S=n+\gamma\,n(n{+}1)/2$.
Because the initial model's attempt $s{+}1$ is distributed as $\mu_0T^{s}$, this proves compression.
For $\gamma=0$, $S=n$ and $\mu_n=\mu_0T^n$.
The $k$-th attempt of the trained model is $\mu_nK_n^{k-1}=\mu_nT^{(n+1)(k-1)}$, which proves amplification.
For $\gamma=0$, it equals $\mu_0T^{(n+1)k-1}$, the initial model's attempt $(n{+}1)k$.

\emph{Fixed point.}
Let $\dob_m=\dob(T^m)<1$.
The update of $K$ alone is the map $K\mapsto KT$, whose $m$-fold iterate $K\mapsto KT^m$ is a $\dob_m$-contraction in $\max_a\TV{K(a,\cdot)-K'(a,\cdot)}$, so $K_n\to K^\star=\one\pinf$, the unique solution of $K^\star=K^\star T$.
Given $K^\star$, a fixed point of the update of $\mu$ satisfies $\mu^\star=(1-\gamma)\mu^\star T+\gamma\mu^\star\one\pinf T=(1-\gamma)\mu^\star T+\gamma\pinf$.
If $\gamma=0$, this is $\mu^\star=\mu^\star T$, whose unique solution is $\pinf$.
If $\gamma>0$, the matrix $I-(1-\gamma)T$ is invertible, because $(1-\gamma)T$ has spectral radius $(1-\gamma)<1$, so $\mu^\star=\gamma\pinf(I-(1-\gamma)T)^{-1}$ is unique.
Since $\pinf(I-(1-\gamma)T)=\gamma\pinf$, this solution is $\pinf$, and it satisfies $\mu^\star=\mu^\star T$.

\emph{Rates.}
Every exponent in the mixture $\mu_n=\E[\mu_0T^S]$ satisfies $S\ge n$.
Hence
\[
\begin{aligned}
\TV{\mu_n-\pinf}
&\le\max_{s\ge n}\dob(T^s)\,\TV{\mu_0-\pinf}\\
&\le\dob_m^{\lfloor n/m\rfloor}\TV{\mu_0-\pinf}.
\end{aligned}
\]
Likewise, $\max_a\TV{K_n(a,\cdot)-\pinf}\le\dob_m^{\lfloor(n+1)/m\rfloor}$.
\end{proof}

\begin{corollary}
\label{cor:helps}
For $n\ge1$ exact distributional backups and the fix/break model of \cref{prop:dyn}, \iwrsi improves the first attempt if and only if $f(1-\acc_1)>b\,\acc_1$.
This is the same condition under which \icrsi helps the initial model (\cref{prop:dyn}).
\end{corollary}

\begin{proof}[Proof of \cref{cor:helps}]
By \cref{thm:main,prop:accelerated}, $\mu_n$ is a mixture of initial-model attempts with index at least $n{+}1$.
By \cref{prop:dyn} with $K=T$, every such attempt is more accurate than the first exactly when $f(1-\acc_1)>b\,\acc_1$, and so is their mixture.
\end{proof}

\paragraph{Accuracy effect of Accelerated Distillation.}
In the fix/break model of \cref{prop:dyn}, let $\lambda=1-f-b$ and $\acc_\infty=f/(f{+}b)$.
After $n\ge1$ exact backups of \cref{def:backup},
\begin{align}
\acc(\mu_n)-\acc_\infty
&=\big(\acc(\mu_0)-\acc_\infty\big)\lambda^n\nonumber\\
&\quad{}\times\prod_{i=1}^n\big((1-\gamma)+\gamma\lambda^i\big).
\label{eq:accelerated-accuracy}
\end{align}
The product quantifies how Accelerated Distillation contracts the remaining accuracy gap beyond Recursive Distillation.

To see this, use the mixture representation $\mu_n=\E[\mu_0T^S]$ from \cref{eq:accelerated-closed}.
By \cref{prop:dyn}, $\acc(\mu_0T^s)-\acc_\infty=(\acc(\mu_0)-\acc_\infty)\lambda^s$.
Linearity of accuracy and independence of the increments $s_i$ give
\begin{align*}
\E[\lambda^S]
&=\prod_{i=0}^{n-1}\big((1-\gamma)\lambda+\gamma\lambda^{i+2}\big)\\
&=\lambda^n\prod_{i=1}^n\big((1-\gamma)+\gamma\lambda^i\big),
\end{align*}
which proves \cref{eq:accelerated-accuracy}.
Since $0\le\lambda<1$ and $0\le\gamma<1$, every factor in the final product is at most one.
When $f(1-\acc_1)>b\,\acc_1$, Accelerated Distillation therefore improves or matches the first-attempt accuracy of the Recursive Distillation backup.

\paragraph{Notes on \cref{thm:main}.}
Without Accelerated Distillation ($\gamma=0$), refinement distillation does not enter the update of $\mu$.
Training answer distillation alone already gives $\mu_n=\mu_0T^n$, and the $k$-th attempt is then the initial model's attempt $n{+}k$.
Training refinement distillation alone instead gives the initial model's attempt $(n{+}1)(k{-}1){+}1$, so exact backups leave the first attempt unchanged.
In practice it still raises the first attempt (\cref{tab:ablation}a), because the two behaviors share parameters and the implemented token-level objective is not an exact distributional backup.
Accelerated Distillation couples the first attempt to the improvement step by making its target $\mu K_nT$ depend on $K_n$.
The distributional convergence rate measures mixing over entire solutions, whereas the fix/break model tracks correctness through the factor $1{-}f{-}b$.
Under the improvement condition of \cref{cor:helps}, the combined backup improves the first attempt, while further refinements approach the same limit.
As $\mu_n$ and every row of $K_n$ approach $\pinf$, the additional gain from in-context improvement vanishes.

\paragraph{Additional adjacent supervision.}
\begin{proposition}
\label{prop:deeper}
Adding adjacent pairs in which a teacher that sees $a_{k+1}$ teaches a student that sees $a_k$, with $k\ge2$ and positive loss coefficients, leaves the exact distributional backup $\Phi$ in \cref{def:backup} unchanged.
\end{proposition}
\begin{proof}
For a student context $a$, every adjacent-pair term averages $\KL(T(a',\cdot)\|K(a,\cdot))$ over $a'\sim K_\theta(a,\cdot)$.
The chain position changes only the sampling weight of row $a$.
By \cref{lem:mixture}, these terms share the minimizer $(K_\theta T)(a,\cdot)$, so the full backup still updates $K$ to $KT$.
The added terms do not involve the first-attempt distribution $\mu$, whose update is therefore unchanged.
\end{proof}
Additional pairs repeat the same refinement target while requiring longer reference chains.
We retain the first two adjacent pairs and compare a third in \cref{tab:ablation}.

\paragraph{Self-Consistency and self-improvement.}
Methods based on majority-conditioned supervision \citep{zuo2025ttrl,gkountouras2026canon} or probability sharpening \citep{agarwal2025entropy,prabhudesai2026rent} favor answers that already dominate the model's predictions, which can reinforce an incorrect majority.
Training on successive reference solutions instead gives the refinement backup, which approaches $\pinf$.
Its accuracy depends on the correction and preservation behavior of $T$ rather than the initial mode.
The comparison on questions with an incorrect initial majority in \cref{tab:axes}a examines this distinction empirically.

\section{Prompts}
\label{app:prompts}

\paragraph{Question-only prompt.}
Evaluation and the first attempt use the plain question prompt:
\begin{promptbox}
\{question\}\\[2pt]
Please reason step by step, and put your final answer within \textbackslash boxed\{\}.
\end{promptbox}

\paragraph{Improvement template.}
The improvement step of \cref{sec:operator} uses:
\begin{promptbox}
\{question\}\\[2pt]
A previous attempt at solving this problem was:\\[2pt]
\{previous solution\}\\[2pt]
Please solve the problem independently. Reason step by step, and put your final answer within \textbackslash boxed\{\}.
\end{promptbox}
The reference $a_k$ is the previous visible solution, including its reasoning process and final answer.
For reasoning models, $a_k$ excludes the hidden thinking trace, which is not carried forward.

\section{Implementation details}
\label{app:impl}

\iwrsi uses LoRA \citep{hu2022lora} and a vLLM rollout engine shared by all ranks \citep{kwon2023vllm}.
\Cref{tab:hparams} lists the hyperparameters.

\begin{algobox}[H]
\centering
\fbox{\begin{minipage}{0.95\linewidth}
\small
\textbf{Input:} model $\pi_\theta$, teacher $\pi_{\theta_0}$, and a batch $\mathcal{Q}$ of unlabeled questions.\\[3pt]
\textbf{for} each $q\in\mathcal{Q}$ \textbf{do}\\
\hspace*{1em}sample $a_1\sim\pi_\theta(\cdot\mid q)$ and $a_2\sim\pi_\theta(\cdot\mid q,a_1)$\\
\hspace*{1em}sample student rollouts $y_0$ given $q$ and $y_1$ given $(q,a_1)$\\
\hspace*{1em}compute $L_{\mathrm{Rec}}$ using \cref{eq:recursive}\\
\hspace*{1em}compute $\omega$ using \cref{eq:omega} and $L_{\mathrm{Acc}}$ using \cref{eq:accelerated}\\
\hspace*{1em}$L_q\gets L_{\mathrm{Rec}}+w_{\mathrm{Acc}}L_{\mathrm{Acc}}$\\
\textbf{end for}\\
Update $\theta$ to minimize $\frac{1}{|\mathcal{Q}|}\sum_{q\in\mathcal{Q}}L_q$.
\end{minipage}}
\caption{\textbf{One update of \iwrsi.}}
\label{alg:iwrsi}
\end{algobox}

\begin{table}[htbp]
\centering
\small
\setlength{\tabcolsep}{3pt}
\begin{adjustbox}{max width=\linewidth}
\begin{tabular}{@{}ll@{}}
\toprule
LoRA & rank 64, scaling 128 \\
Optimizer & AdamW \\
Learning rate & $5{\times}10^{-6}$ \\
Updates & 100 \\
Batch & 32 question occurrences / update \\
Precision & bf16 \\
Seed & 42 \\
\midrule
Teacher & frozen initial model \\
OPSD forward KL & temp.\ 1.1, per-token clip 0.05 \\
Loss weights & 1 for $L_{\mathrm{ans}}$, $L_{\mathrm{ref}}$, $L_{\mathrm{Acc}}$ \\
Token weights & $\alpha{=}0.5$, $\tau{=}10^{-4}$ \\
\midrule
Reference mode & reasoning \\
Reference sampling & temp.\ 1.0, top-$p$ 0.95 \\
Generation / context & 38{,}912 / 40{,}960 tokens \\
Student rollouts & non-thinking, up to 1{,}024 tokens \\
\bottomrule
\end{tabular}
\end{adjustbox}
\caption{\textbf{Core hyperparameters of \iwrsi.}}
\label{tab:hparams}
\end{table}

\paragraph{Pipelined generation.}
\label{app:pipeline}
The pipeline batches the current update's $a_2$ and student rollouts with the next update's $a_1$, so the prefetched $a_1$ is one update older.
All \iwrsi results use this schedule.

\paragraph{Evaluation sampling.}
\label{app:eval}
All evaluations use reasoning (thinking) mode with temperature 1.0, top-$p$ 0.95, top-$k$ disabled ($-1$), min-$p$ 0 and presence penalty 0.
Each response may generate up to 38{,}912 tokens within a 40{,}960-token context, and we draw 12 samples per question (64 for \cref{tab:axes}a).

\section{Baseline details}
\label{app:baselines}

\paragraph{TTPO.}
Following the released implementation \citep{wang2026ttpo}, TTPO selects eight training trajectories from 64 non-thinking candidates per question.
A teacher conditioned on the majority answer supervises agreeing trajectories through Soft-OR-weighted forward KL, while a group-relative policy-gradient term penalizes disagreements.

\paragraph{TTRL.}
TTRL optimizes the policy with GRPO using agreement with the sampled majority as its reward \citep{zuo2025ttrl}.
In the training-cost comparison, it generates eight thinking trajectories per question and reuses each batch for two updates.

\paragraph{OPSD-TTT.}
OPSD-TTT uses the current policy's generated final answer to condition a frozen initial-model teacher \citep{wang2026ttpo}.
The teacher supervises one non-thinking student rollout per question through clipped forward KL.
LoRA and optimizer settings follow \cref{tab:hparams}.

\paragraph{CANON.}
CANON samples 32 solutions per question and selects the solution with the highest mean token log-probability among those reaching the majority answer \citep{gkountouras2026canon}.
This full solution conditions a frozen teacher, and the question-only student is trained with Jensen--Shannon divergence over the sampled rollouts.

\paragraph{U-OPSD.}
U-OPSD samples eight rollouts per question and uses a majority-agreeing solution as the teacher's reference to supervise disagreeing rollouts through clipped forward KL \citep{uopsd2026}.
The training-cost comparison uses thinking mode for generation and distillation.

\paragraph{Comparison setup.}
All baselines use the same starting checkpoints and unlabeled benchmark questions as \iwrsi, with final boxed answers graded by math-verify.

The training-cost comparison in \cref{tab:training-efficiency}b uses Qwen3-4B on four H20 GPUs, with 100 updates and 32 question occurrences per update.
As in all other experiments, \iwrsi uses pipelined references and $\tau=10^{-4}$.

\paragraph{Test-time scaling budget.}
\label{app:budget}
Under the same test-time scaling budget, TTPO + SC generates as many tokens per question as \ttrsi uses to reach its final attempt, i.e., the initial solution and ten refinements.
\icrsi also reads the previous solution at each step.
Because $a_k$ excludes the hidden thinking trace (\cref{app:prompts}), it averages fewer than 1{,}000 tokens, far fewer than those generated per attempt (\cref{fig:tts}).
We therefore count only generated tokens and treat this input cost as negligible.

\section{Case studies}
\label{app:case-studies}

We verify the recurrence in \cref{fig:case-studies} and add a geometric example that traces corrections to the calculation and the final conclusion.

\paragraph{Stick breaking.}
Let $E(n)$ be the expected length of the piece containing the red endpoint after $n$ splits.
There are then $n+1$ pieces, and this piece is halved with probability $1/(n+1)$.
Thus
\[
\begin{aligned}
E(n+1)&=E(n)\left(1-\frac{1}{2(n+1)}\right)\\
&=E(n)\frac{2n+1}{2(n+1)}.
\end{aligned}
\]
The expected sum of squared piece lengths $S(n)$ follows the same recurrence, since a split halves the piece's contribution.
Thus $S(n)=E(n)$ from $S(0)=E(0)=1$, giving $63/256$ after five splits.
\paragraph{Geometry.}
The base chain in \cref{fig:case-geometry} corrects its coordinates at $a_{10}$ while still reporting the old answer, and only updates the answer at $a_{11}$.
This illustrates how intermediate reasoning can become correct a step before the stated answer does.
After \iwrsi training, $a_5$ replaces a guessed angle with coordinates and a cosine-rule check.

With $A=(0,0)$, $B=(1,0)$ and $C=(\cos92^\circ,\sin92^\circ)$, $BP^2=PQ^2=QC^2$ gives $AP\approx0.5817$, $AQ\approx0.6687$ and $\angle APC\approx74^\circ$.

\begin{figure}[H]
\centering
\setlength{\abovedisplayskip}{3pt}
\setlength{\belowdisplayskip}{3pt}
\setlength{\abovedisplayshortskip}{2pt}
\setlength{\belowdisplayshortskip}{2pt}
\input{figures/case_geometry}
\vspace{10pt}
\caption{\textbf{Refining a geometric solution.}
Selected HMMT26 reasoning before and after \iwrsi training, with blue marking revisions.}
\label{fig:case-geometry}
\end{figure}

\end{document}